\documentclass[pdflatex,sn-basic]{sn-jnl}

\usepackage{tabularx}

\jyear{2021}%

\theoremstyle{thmstyleone}%
\theoremstyle{thmstyletwo}%

\theoremstyle{thmstylethree}%

\begin{document}

\title[Convolutional Neural Networks combined with Runge--Kutta Methods]{Convolutional Neural Networks combined with Runge--Kutta Methods}


\author[1]{\fnm{Mai} \sur{Zhu}}\email{zhumai@stumail.neu.edu.cn}

\author[2]{\fnm{Bo} \sur{Chang}}\email{bchang@stat.ubc.ca}

\author*[1]{\fnm{Chong} \sur{Fu}}\email{fuchong@mail.neu.edu.cn}

\affil*[1]{\orgdiv{School of Computer Science and Engineering}, \orgname{Northeastern University}, \orgaddress{
\city{Shenyang},
\state{Liaoning}, \country{China}}}

\affil[2]{\orgdiv{Department of Statistics}, \orgname{University of British Columbia}, \orgaddress{
\city{Vancouver},
\country{Canada}}}


\abstract{A convolutional neural network can be constructed using numerical methods for solving dynamical systems, since the forward pass of the network can be regarded as a trajectory of a dynamical system. However, existing models based on numerical solvers cannot avoid the iterations of implicit methods, which makes the models inefficient at inference time. In this paper, we reinterpret the pre-activation Residual Networks (ResNets) and their variants from the dynamical systems view. We consider that the iterations of implicit Runge--Kutta methods are fused into the training of these models. Moreover, we propose a novel approach to constructing network models based on high-order Runge--Kutta methods in order to achieve higher efficiency. Our proposed models are referred to as the Runge--Kutta Convolutional Neural Networks (RKCNNs). The RKCNNs are evaluated on multiple benchmark datasets. The experimental results show that RKCNNs are vastly superior to other dynamical system network models: they achieve higher accuracy with much fewer resources. They also expand the family of network models based on numerical methods for dynamical systems.}




\maketitle

\section{Introduction}\label{sec:intro}

The neural network community has long been aware of the numerical methods for dynamical systems. The Runge--Kutta Neural Network (RKNN) is proposed for the identification of unknown time-invariant dynamical systems by \cite{wang1998runge}. RKNNs conform exactly to the formula of the Runge--Kutta (RK) methods; i.e. the specific time-step size and the precise coefficients of the RK methods. In RKNNs, a neural network is used to approximate the Ordinary Differential Equation (ODE), which governs the rate at which the system states change. Adopting the RK methods brings higher prediction accuracy and better generalization capability into the neural network \citep{wang1998runge}. However, it has not been used to model the visual system or extended to the Convolutional Neural Networks (CNNs).

Recently, \cite{chen2018neural} proposed the RK-Nets and ODE-Nets. The RK-Nets can be regarded as the time-variant convolutional version of RKNNs, while the ODE-Nets can be considered as extending the RK-Nets from the RK methods to the linear multi-step (LM) methods. Moreover, \cite{chen2018neural} evaluate the RK-Net and ODE-Net on the Modified National Institute of Standards and Technology (MNIST) dataset\footnote{\url{http://yann.lecun.com/exdb/mnist/}}, an image classification dataset. As a series of derived models, the RK-Nets and ODE-Nets use RK and LM methods just like their ancestors, the RKNNs. To be specific, the neural network is only used to approximate the ODE during the process of a numerical approximation.

A review of existing approaches often reveals some problems, like the work done by \cite{lu2020few, GAVAHI2021115511,morales2021playing}. According to \cite{endre2003an}, both RK and LM methods could be explicit or implicit. Explicit methods calculate the current state of a dynamical system from the previous state of the system. By contrast, implicit methods need to solve an equation that involves both the previous and current state. The common approach for solving the equation of implicit methods is to approximate it with a sequence of iterations. These iterations are needed both during training and inference of RKNNs, RK-Nets, and ODE-Nets, since the approximation is independent from the neural network. Hence, the models mentioned above are inefficient. The most direct way to improve their efficiencies is to modify the architecture of the neural network for ODE. Nevertheless, the iterations for implicit equations, which cost much computation time and memory, are always present regardless of the choice of the neural network for ODE, and degrade the performance severely. Consequently, it is an important research question to search for other ways to construct more efficient numerical network models.

We focus on utilizing the RK methods to construct network models, since RK methods are usually the building blocks of LM methods. In a time-step, RK methods calculate the derivatives in several stages from the ODE and then use the weighted average of these derivatives as the estimated rate of change of the system states. The RK methods have two families, the explicit RK (ERK) methods and the implicit RK (IRK) methods. The IRK methods are more stable and have a higher order than the ERK methods with the same stages \citep{butcher2008numerical}. Since the higher-order RK methods have lower truncation error, the classification accuracy is able to be enhanced by adopting them. Thus, we utilize the IRK methods to construct the network models. However, the existing implementation of IRK methods uses a Newton method, which is a sequence of iterations to converge to the acceptable value. This is a process of approximating the equation of IRK methods. It could be approximated by a neural network due to its versatility in approximation. Therefore, we try to combine the approximation of IRK equation and the neural network for ODE in order to utilize the IRK methods efficiently.

In the past few years, researchers have studied the relation between ResNets and dynamical systems \citep{liao2017towards,E2017,haber2018learning,chang2018reversible,chang2018multi,pmlr-v80-lu18d}. ResNets are feed-forward CNNs with a skip connection \citep{He_2016_CVPR}. They have achieved great success on several vision benchmarks \citep{He_2016_CVPR}. The forward Euler method, a first-order ERK method, has been employed to explain the ResNets with full pre-activation \citep{he2016identity} from a dynamical systems point of view \citep{haber2018learning,chang2018multi}. Nevertheless, there is no firm evidence that the residual block is just the forward Euler method but not any other RK method. The local truncation error of the residual block is impossible to be fixed in the first order since the accuracy of neural networks is variable under different conditions such as input, training, etc.

By contrast, we reinterpret the residual mapping in a residual block as an approximation to the increment in a time step without any detail of some RK method. The equation of RK methods, including its coefficients, is approximated as a whole. Moreover, the accuracy of the approximation is determined by the structure of CNN and the training. In other words, the pre-activation ResNet and its variants, which focus on improving the residual mapping, do not correspond to the forward Euler method exactly but to RK methods. The approximated RK methods can be implicit due to the versatility of neural networks on approximation. Hence, our new explanation provides a feasible approach implementing the IRK methods within the network structure. In other words, the independent iterations for approximating IRK equation, which cost much time and memory, are eliminated. All approximations are contained within the neural network itself. Thus, for efficiency improvement, the improvement of the neural network would play a bigger role than in RKNNs, RK-Nets, and ODE-Nets when the implicit methods are used. Next, we introduce how to improve the neural network of RK methods.

For the performance, we consider that the lack of details of the RK methods in the pre-activation ResNets is as bad as the excessive details in RKNNs and RK-Nets. Hence, we improve the residual mapping by expressing moderate details of the RK methods. To be specific, the pre-activation ResNets approximate the increment in each time-step as a whole, which is the product of the time-step size and the weighted average of derivatives at all stages in the RK equation. On the other hand, RKNNs and RK-Nets approximate the ODE, which is used to calculate the derivative in each stage. Nevertheless, we use the subnetwork to approximate the increment in each stage of a time-step, which is the product of the time-step size, quadrature weight, and derivative in each stage. As a result, we propose a novel and efficient network architecture adopting the RK methods, called RKCNN.

We evaluate the performance of RKCNNs on the benchmark datasets, including MNIST, the Street View House Numbers (SVHN) dataset \citep{37648}, and the Canadian Institute for Advanced Research (CIFAR) dataset \citep{krizhevsky2009learning}. The experimental results show that RKCNNs are much more efficient than the state-of-the-art (SoTA) network models related to the numerical methods on these datasets.

In summary, the main contributions of our work are:
\begin{itemize}
\item We provide a new explanation for the pre-activation ResNet and its variants which focus on improving the residual mapping. We consider that these models adopt the RK methods and not only the forward Euler method. Thus, we offer a new direction of thinking on the network structure. In this approach, the burdensome iterations of IRK methods are eliminated.
\item We propose a novel and efficient neural network architecture inspired by the RK methods, which is called RKCNN.
In RKCNNs, the neural network of a time-step consists of an identity mapping of the initial state of this step and several subnetworks for stages. Each stage is approximated by convolutional layers under the rules of RK methods.
We enrich the family of network models based on numerical methods.
\end{itemize}

The rest of the paper is organized as follows. The related work is reviewed in Sect. \ref{sec:related_work}. The architecture of RKCNNs is described in Sect. \ref{sec:RKCNNs}. The performance of RKCNNs is evaluated in Sect. \ref{sec:experiments}. The conclusion and the future work are described in Sect. \ref{sec:conclusion}.

\section{Related work}\label{sec:related_work}

The RK methods are commonly used to solve ODEs in numerical analysis. The forward Euler method is a first-order RK method. Higher-order RK methods can achieve lower truncation errors than lower-order RK methods, including the forward Euler method. Moreover, RK methods are usually the building blocks of LM methods. Therefore, the RK methods are ideal tools to construct network models from the dynamical systems view.

The RK methods have been adopted to construct neural networks, which are known as RKNNs, for the identification of the unknown time-invariant dynamical systems described by ODEs. Neural networks are classified into two categories \citep{wang1998runge}: (\romannumeral1) a network that directly learns the state trajectory of a dynamical system, called a direct-mapping neural network (DMNN); (\romannumeral2) a network that learns the rate of change of the system states, called RKNN.
RKNNs are proposed to eliminate several drawbacks of DMNNs, such as the difficulty in obtaining high accuracy for the multi-step prediction of the state trajectories. It has been shown theoretically and experimentally that the RKNN has higher prediction accuracy and better generalization capability than the conventional DMNN.

Recently, \cite{chen2018neural} proposed the RK-Nets and ODE-Nets. These network models use a convolutional subnetwork to approximate an ODE, similar to the RKNNs. However, they explicitly deal with the time variable in the subnetwork in order to support the time-variant system. RKNNs, RK-Nets, and ODE-Nets implement the numerical methods following common approaches in mathematics. For the implicit methods, they require iterations to converge to acceptable accuracy. Hence, they cost more time and memory than the explicit methods. For the explicit methods, RKNNs, RK-Nets, and ODE-Nets are not efficient in image classification. They are even less efficient than the pre-activation ResNets, which are the base network adopted by \cite{chen2018neural}.

On the other hand, some work has emerged to connect dynamical systems with deep neural networks \citep{E2017}, or in particular ResNets \citep{haber2018learning,chang2018reversible,chang2018multi,pmlr-v80-lu18d,JMLR:v18:17-653}. ResNets are deep feed-forward networks with identity mappings as shortcuts. They have gained much attention over the past few years since they have obtained impressive performance on many challenging image tasks, including in medical fields \citep{OZTURK2021102601,ozturk2021attention}. \cite{liao2017towards} regards ResNet with pre-activation as an unfolded shallow recurrent neural network which implements a discrete dynamical system. This provides a novel point of view for understanding the pre-activation ResNets from the dynamical systems view. \cite{E2017} proposes to use continuous dynamical systems as a tool for machine learning and interprets the residual block in the pre-activation ResNets as a discretization of the dynamical system. \cite{haber2018learning} interpret this residual block as a forward Euler discretization.

Based on the same interpretation as \cite{haber2018learning}, the following works emerge. \cite{chang2018reversible} propose three reversible architectures with order 2, based on ResNets and ODE systems. \cite{chang2018multi} propose a novel method for accelerating ResNets training. \cite{JMLR:v18:17-653} presented a training algorithm that can be used in the context of ResNets. \cite{pmlr-v80-lu18d} propose a 2-step architecture based on LM methods and regard the midpoint and leapfrog network structures of \cite{chang2018reversible} as their special cases. \cite{chen2018neural} work is also based on this interpretation. It adds the time variable in the residual mapping of ResNets and uses this transformed subnetwork to approximate the ODE in RK-Nets and ODE-Nets. ODE-Nets \citep{chen2018neural} extend the application of LM methods from 2-step methods to more multistep methods with higher orders. \cite{NEURIPS2019_21be9a4b} augment the space on which the ODE is solved based on RK-Nets and ODE-Nets. \cite{NEURIPS2020_418db2ea} extend \cite{NEURIPS2019_21be9a4b} from first-order ODEs to second-order ODEs. \cite{pmlr-v139-sander21a} reduce the memory requirement of ResNets and interpret the proposed Momentum ResNets as second-order ODEs. In these references, the residual mapping of ResNets is regarded as the increment in a time-step of the forward Euler method. To construct efficient models, we focus on the improvement of the residual mapping since we reinterpret it as approximating some RK methods.

We approximate the IRK methods in the network structure together with the ERK methods. The subnetwork of each time step is trained to implement the RK methods. To construct this subnetwork, we use the dense block of a Dense Convolutional Network (DenseNet) \citep{Huang_2017_CVPR} and the clique block of a convolutional neural network with alternately updated clique (CliqueNet) \citep{Yang_2018_CVPR} for reference according to the transformation of the equations.

DenseNets are state-of-the-art network models extending ResNets. The dense connection is the main difference between them. There are direct connections from a layer to all the subsequent layers in a dense block in order to allow better information and gradient flow. CliqueNets are state-of-the-art network models based on DenseNets. They adopt the alternately updated clique blocks to incorporate both forward and backward connections between any two layers in the same block. Our RKCNNs not only surpass the numerical network models but also exceed DenseNets and CliqueNets.

\section{RKCNNs}
\label{sec:RKCNNs}

We provide an overview of the RK methods in Sect. \ref{subsec:RKmethod}. The overall structure of RKCNNs is described in Sect. \ref{subsec:RKCNN}. We elaborated on the structure of the subnetwork for increment in each time step in Sect. \ref{subsec:subnet}.

\subsection{Runge--Kutta methods}
\label{subsec:RKmethod}

An initial value problem for a time-dependent first-order dynamical system can be described by the following ODE \citep{butcher2008numerical}:
\begin{equation}
 \frac{dy} {d t} =f\left( t,\ y(t)\right),\qquad y\left( t_{0}\right) = y_{0},\label{eq:ode}
\end{equation}
where $y$ is a vector representing the system state. The dimension of $y$ should be equal to the dimension of the dynamical system. The ODE in Eq \eqref{eq:ode} represents the rate of change of the system states. The rate of change is a function of time $t$ and the current system state $y(t)$. RK methods utilize the rate of change calculated from the ODE to approximate the increment in each time step, and then obtain the predicted final state at the end of each step. The RK methods are numerical methods originated from the Euler method. There are two types of RK methods: explicit and implicit. Both of them are employed in RKCNNs. The family of RK methods is given by the following equations \citep{endre2003an}:
\begin{equation}
y_{n+1} = y_{n} + h\sum^{s}_{i=1}b_{i} z_{i},\qquad t_{n+1}=t_{n}+h,\label{eq:add}
\end{equation}
where
\begin{equation}
 z_{i} = f\left( t_{n}+c_{i}h,\  y_{n}+h\sum^{s}_{j=1}a_{ij} z_{j}\right),\qquad 1\leq{i}\leq{s}.\label{eq:general}
\end{equation}
In Eq \eqref{eq:add}, $y_{n+1}$ is an approximation of the solution to Eq \eqref{eq:ode} at time $t_{n+1}$, i.e. $y(t_{n+1})$; $y_0$ is the input initial value; $h\sum^{s}_{i=1}b_{i}z_{i}$ is the increment of system state $y$ from $t_n$ to $t_{n+1}$; $\sum^{s}_{i=1}b_{i}z_{i}$ is the estimated slope, which is a weighted average of the slopes $z_i$ computed in different stages. The positive integer $s$ is the number of $z_i$, i.e. the number of stages of the RK method. Eq \eqref{eq:general} is the general formula of $z_i$. $h$ is the time-step size that can be adaptive for different time steps.

In numerical analysis, $s$, $a_{ij}$, $b_i$, and $c_i$ in Eq \eqref{eq:add} and Eq \eqref{eq:general} need to be prespecified for a particular RK method. These coefficients are displayed in a partitioned tableau \citep{butcher2008numerical}. The ERK is methods with $a_{ij} = 0$ when $1\leq{i}\leq{j}\leq{s}$. All the RK methods other than ERK are IRK methods. The algebraic relationship of the coefficients has to meet the order conditions to reach the highest possible order. Different RK methods have different truncation errors, which are denoted by the order: an order $p$ indicates that the local truncation error is $O(h^{p+1})$. If an $s$-stage ERK method has order $p$, then $s\geq{p}$; if $p\geq 5$, then $s>p$ \citep{butcher2008numerical}. Furthermore, an $s$-stage IRK method can have order $p = 2s$ when its coefficients are chosen under some conditions.
Therefore, more stages may achieve higher orders, i.e. lower truncation errors. The Euler method is a one-stage first-order RK method with $b_1=1$ and $c_1=0$.
In other words, the high-order RK methods can be expected to achieve lower truncation errors than the Euler method. Thus, the goal of our proposed RKCNNs is to improve the classification accuracy by taking advantage of the high-order RK methods.

It is necessary to specify $h$ in order to control the error of approximation in common numerical analysis. The varying time-step size can be adaptive to the regions with the different rates of change. The truncation error is lower when $h$ is smaller.

\subsection{From RK methods to RKCNNs}\label{subsec:RKCNN}

\begin{figure*}[h]
\centering
\includegraphics[width=\textwidth]{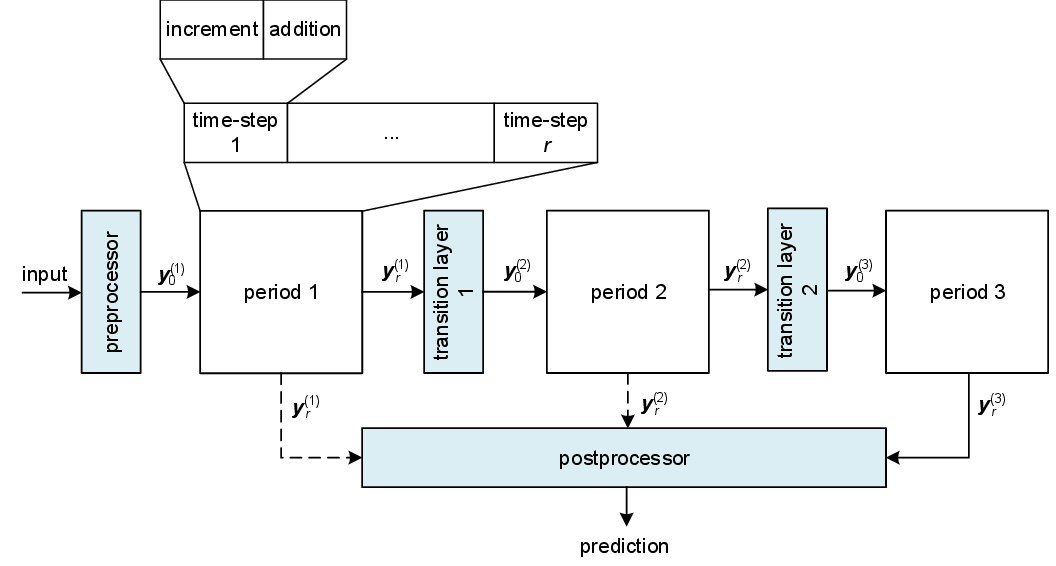}
\caption{Architecture of a 3-period RKCNN. $y^{(d)}$ denotes the system state of period $d$. $y_0^{(d)}$ is the initial state of period $d$. $y_r^{(d)}$ is the final state after $r$ time-steps in period $d$. $r$ is the total number of time steps in a period. It can vary in different periods. Period 1 and time-step 1 in it are unfolded as an example. System state changes throughout a period. The final state of a step is estimated as the initial state of this step, adding an increment. This operation originates from the RK methods. The approximation of the increment is the key point in RKCNNs. The dotted lines show the multiscale feature strategy.}
\label{fig:arch}
\end{figure*}

There are three components in RKCNNs: the preprocessor, the multi-periods, and the postprocessor. The preprocessor manipulates the raw images and passes the results to the first period. The postprocessor deals with the output from the last period and then passes the result to the classifier to make a decision. The periods between those two components are divided by the transition layers. These periods can be modeled by the time-dependent dynamical systems. Each period of an RKCNN is divided into $r$ time steps as shown in \figurename~\ref{fig:arch}. The RK methods approximate the final state of every time step using the rate of change of the system states.
Each period can be written as below.
\begin{equation}
y_r = B(\rho,\ y_0)
\end{equation}
Here, $B(\cdot)$ is the convolutional network approximating the period. $\rho$ is the network parameters. $y_0$ is the input of the network and also the initial state of the period. $y_r$ is the output of the network and also the final state of the period. $B(\cdot)$ consists of $r$ subnetworks connected end to end, each using the RK method to approximate a time step. Some guiding principles when applying the RK methods to RKCNNs are listed as follows.

Firstly, dimensionality reduction is often carried out to simplify the system identification problem when the dimensions of the real dynamical system are too high. The dimension of each period in RKCNNs, i.e. the dimension of $y$ in Eq \eqref{eq:ode}, is predefined as the product of the feature map size and the number of channels at the beginning of a period. The dimensions of $y$ in the same periods of different RKCNNs can be different due to various degrees of dimensionality reduction. Nevertheless, the dimension of $y$ is consistent within a period.

Secondly, given that there is no explicit ODE for image classification, a convolutional subnetwork is employed to approximate the increment in each time step. The number of units in each hidden layer of this subnetwork can be more than the dimension of $y$.

Thirdly, the number of stages $s$ in each period is predefined in RKCNNs, but the other coefficients, $a_{ij}$, $b_i$, and $c_i$ in Eq \eqref{eq:add} and Eq \eqref{eq:general}, are learned by back-propagation. Due to the order conditions \citep{butcher2008numerical}, the functional relationship among the coefficients is more important than the specific value of any individual coefficient. The optimal relationship among the learned coefficients with the highest possible order is obtained after training. Whether the coefficients are learned implicitly or explicitly does not affect the relationship among them. In order to be efficient, the coefficients are learned implicitly in RKCNNs.

Lastly, the number of time-steps $r$ in each period is predefined in RKCNN, but the step size $h$ is learned by training. $n$ in Eq \eqref{eq:add} and Eq \eqref{eq:general} is limited to the range of [0, $r$). In theory, the adaptive time-step size can achieve higher accuracy. Therefore, different time steps learn their own $h$ separately. For the purpose of classifying images, the specific value of $h$ of each step is not relevant. Thus, $h$ is learned implicitly in RKCNNs for efficiency.

In an RKCNN, a variety of RK methods can be adopted in different periods, while the stages of the RK methods are fixed within one period. The models are named after the specific method in each period, such as RKCNN-3\_4\_2. The suffix in the name of an RKCNN is composed of several $s$ terms; each stands for the number of stages of the RK methods in the corresponding period. The number of such terms equals the total number of periods. $s$ can vary in the different periods. For example, RKCNN-3\_4\_2 has three periods that adopt the 3-stage RK methods, the 4-stage RK methods, and the 2-stage RK methods, respectively. We use this notation throughout this paper.

Given an RKCNN model, $s$ and $r$ can be modified to construct more variants with the same dimensions in the corresponding periods. In other words, $s$ and $r$ control the depth of the network, while dimensionality reduction controls the width of the network. More stages, more time steps, and larger dimensions usually lead to higher classification accuracy. However, the complexity of an ODE increases with increasing dimensions. As a result, the convolutional subnetwork, which approximates the increment in a time step, needs to be more complex for the larger dimensions. Hence, the accuracy is also associated with the matching degree of the dimension and the convolutional subnetwork. Unmatched high-dimensional network models may have lower accuracy. In addition, the training method might affect the classification accuracy too. As mentioned in Sect. \ref{subsec:RKmethod}, the IRK methods can reach higher order than the ERK methods with the same stage. Adequate training and a capable convolutional subnetwork are necessary to learn the IRK methods in RKCNNs. On the contrary, inadequate training or the incompetent convolutional subnetwork may make the functional relationship downgrade from the IRK methods to the ERK methods since the functional relationship of the IRK methods is more complicated than the ERK methods. Consequently, we propose several subnetwork structures to look for the most suitable one. We introduce these structures in the next section.

\subsection{Subnetwork in each time step}\label{subsec:subnet}

We propose the subnetwork structures with the incremental details of the RK methods based on the pre-activation ResNets.
According to \cite{E2017, chang2018reversible}, ResNets use the following formula to approximate the dynamical system.
\begin{equation}
y_{n+1} = y_{n} + hP(y_{n}, \theta_{n}). \label{eq:ewn}
\end{equation}
Here, $\theta_{n}$ is the network parameters of the $n$th residual block, where $y_{n}$ is the input and $y_{n+1}$ is the output. The training of the network will learn $\theta(t)$. Nevertheless, we consider that $\theta$ could be a function of both $t$ and the coefficients of the RK methods due to the versatility of neural networks on approximation. For instance, in $\theta(t, A, b, c)$, $A$ is the matrix of all the $a_{ij}$, while $b$ and $c$ are the vectors of $b_i$ and $c_i$, respectively. Therefore, the pre-activation ResNet and its variants, which focus on improving the residual mapping, can be regarded as special cases of the RK methods.

However, in the pre-activation ResNets, to approximate the increment in a time-step as a whole loses the relations among the stages. On the contrary, the RKNNs and the RK-Nets keep every relation but do not obtain higher performance. We consider that the network model can be more efficient if the relationship among the stages is reflected in the network structure in moderate detail. Hence, we construct RKCNNs in the following way.

Let $e_i$ denote the increment of each stage within a time step. i.e.
\begin{equation}
e_i = hb_{i}z_i. \label{eq:ei}
\end{equation} 
Then, Eq \eqref{eq:add} can be rewritten as below:
\begin{equation}
y_{n+1} = y_{n} + \sum^{s}_{i=1}e_{i}. \label{eq:add_E}
\end{equation} 

On the basis of Eq \eqref{eq:add_E}, we construct the convolutional subnetwork for every time-step in the RKCNNs, called the RK block. In order to construct the RK block, we have to find out the relations among $y_n$ and $e_i$ for $i=1,\ \ldots,\ s$.
Hence, $e_i$ is approximated as below.
\begin{align}
e_i\ &=\ hb_{i}f\left( t_{n}+c_{i}h,\  y_{n}+h\sum\limits^{s}_{j=1}a_{ij} z_{j}\right) \label{eq:D} \\
&=\ D\left(\delta(t_{n}, b_i, c_i),\ y_{n} + h\sum^{i-1}_{j=1}a_{ij}z_{j} + h\sum^{s}_{j=i}a_{ij}z_{j}\right) \label{eq:D2} \\
&=\ M\left(\mu(t_{n}, b_i, c_i),\ y_{n}+h\sum^{s}_{j=i}a_{ij}z_j,\ h\sum^{i-1}_{j=1}a_{ij}z_j \right). \label{eq:M}
\end{align}
Firstly,
$e_i$ is written as Eq \eqref{eq:D} according to Eq \eqref{eq:general} and Eq \eqref{eq:ei}. Therefore, $e_i$ is approximated by a convolutional network $D(\cdot)$ with the parameter $\delta$. $h$ is absorbed into $\delta$ since the adaptive time-step size $h$ is a function of $t$. Moreover,
$h\sum^{s}_{j=1}a_{ij}z_{j}$ is split into two parts, $h\sum^{i-1}_{j=1}a_{ij}z_j$ and $h\sum^{s}_{j=i}a_{ij}z_j$, in Eq \eqref{eq:D2}.
These two parts can be inputted separately. Their summation is approximated by the network.
As a result, the network is transformed into $M(\cdot)$ with the parameter $\mu$ and the inputs $y_n+h\sum^{s}_{j=i}a_{ij}z_j$ and $h\sum^{i-1}_{j=1}a_{ij}z_j$.

Let $A^T = (\alpha_{1},\ldots, \alpha_{s})$. Eq \eqref{eq:general} can be approximated as follows:
\begin{equation}
z_{i} = K(\kappa(t_{n}, \alpha_{i}, c_i),\ y_{n}). \label{eq:K}
\end{equation}
Here, $\kappa$ is the parameters of the network. On the basis of Eq \eqref{eq:K} and \eqref{eq:M}, it can be transformed as follows.
\begin{align}
e_i &=\ M\left(\mu(t_{n}, b_i, c_i),\ y_{n}+h\sum^{s}_{j=i}a_{ij}K(\kappa(t_{n}, \alpha_{i}, c_i),\ y_n),\ h\sum^{i-1}_{j=1}a_{ij}z_j \right) \\
&=\ M\left(\mu(t_{n}, b_i, c_i),\ L(\sigma(t_{n}, \alpha_{i}, c_i),\ y_{n}),\ h\sum\limits^{i-1}_{j=1}a_{ij}z_j \right) \\
&=\ N\left(\psi(t_{n}, \alpha_{i}, b_i, c_i),\ y_{n},\ h\sum^{i-1}_{j=1}a_{ij}z_j \right) \label{eq:N} \\
&=\ N\left(\psi(t_{n}, \alpha_{i}, b_i, c_i),\ y_{n},\ h\sum^{i-1}_{j=1}b_{j}z_j\frac{a_{ij}}{b_j}\right) \label{eq:N2} \\
&=\ E\left(\epsilon(t_{n}, \alpha_{i}, b_i, c_i),\ y_{n},\ hb_{1} z_1,\ \ldots,\ hb_{i-1}z_{i-1}\right) \\
&=\ E\left(\epsilon(t_{n}, \alpha_{i}, b_i, c_i),\ y_{n},\ e_1,\ \ldots,\ e_{i-1}\right).\label{eq:E}
\end{align}
We replace $z_j$ in $y_n+h\sum^{s}_{j=i}a_{ij}z_j$ in Eq \eqref{eq:M} with $K(\cdot)$ according to Eq \eqref{eq:K}.
Thus, $y_n+h\sum^{s}_{j=i}a_{ij}z_j$ can be approximated by a network $L(\cdot)$ with the parameter $\sigma$ and the input $y_{n}$.
Consequently, $e_i$ can be approximated by a network with the parameter $\psi$ and the inputs $y_{n}$ and $h\sum^{i-1}_{j=1}a_{ij}z_j$. This network is written as $N(\cdot)$ in Eq \eqref{eq:N}. Afterwards, every $a_{ij}$ is adjusted to $b_{j}\frac{a_{ij}}{b_j}$ where $b_j \not=0$ in Eq \eqref{eq:N2}.
Each $hb_{j} z_j$ can be inputted separately. Their weighted summation is approximated by the network.
As a result, $e_i$ can be approximated by a network with the parameter $\epsilon$ and the inputs $y_{n}$ and $hb_{j} z_j$ for $j=1,\ \ldots,\ i-1$. This network is written as $E(\cdot)$. After replacing each $hb_{j} z_j$ with $e_j$ according to Eq \eqref{eq:ei}, $e_i$ is approximated by a network with the inputs $y_n$ and $e_j$ for $j=1,\ \ldots,\ i-1$. Eq \eqref{eq:E} reflects the relationship among the increments in partial stages.

We consider that the dense connection in DenseNets is the most similar network structure to approximate Eq \eqref{eq:E}. To be specific, all the preceding layers in a dense block are concatenated as the input of the following subnetwork. It is just like Eq \eqref{eq:E} that uses $y_n$ and all the increments in the preceding stages as its input. Thus, we use the dense block for reference to approximate the increment in each stage. The RKCNN constructed in this way is denoted as RKCNN-E.

To be specific, an RK block is composed of a restricted dense block followed by a summation layer. The input of the RK block is $y_n$. The output of the restricted dense block is $y_n$ and all the increments $e_i$ for $i=1,\ \ldots,\ s$. The summation layer adds $y_n$ and $e_i$ for $i=1,\ \ldots,\ s$ to obtain $y_{n+1}$ according to Eq \eqref{eq:add_E}. The restricted dense block must obey the following rules:

\textbf{Rule a}: The number of channels of $y_n$ is restricted to the growth rate of the dense block. They are both written as $k$.

\textbf{Rule b}:
The total times of growth in the dense block are $s$, which is the number of stages of the RK methods. $k$ channels outputted by the $i$th growth are $e_i$ for $i=1,\ \ldots,\ s$.

Eq \eqref{eq:E} reflects the relationship between any stage and the stages before it.
Nevertheless, we consider that the relationship between any stage and the stages after it can be expressed, too. Therefore, we transform Eq \eqref{eq:D2} into the following form:
\begin{align}
e_i &= D\left(\delta(t_{n}, b_i, c_i),\ y_{n} + h\sum^{i-1}_{j=1}a_{ij}z_{j} + ha_{ii}z_{i} + h\sum^{s}_{j=i+1}a_{ij}z_{j}\right) \label{eq:D3} \\
&= G\left(\beta(t_{n}, b_i, c_i),\ y_{n}+ha_{ii}z_{i},\ h\sum^{i-1}_{j=1}a_{ij}z_j,\ h\sum^{s}_{j=i+1}a_{ij}z_j\right) \\
&= G\left(\beta(t_{n}, b_i, c_i),\ y_{n}+ha_{ii}K(\kappa(t_{n}, \alpha_{i}, c_i),\ y_n),\ \right. \nonumber \\
&\qquad\qquad\left. h\sum^{i-1}_{j=1}a_{ij}z_j,\ h\sum^{s}_{j=i+1}a_{ij}z_j\right) \\
&= G\left(\beta(t_{n}, b_i, c_i),\ J(\upsilon(t_{n}, \alpha_{i}, c_i),\ y_n),\  h\sum\limits^{i-1}_{j=1}a_{ij}z_j,\ h\sum\limits^{s}_{j=i+1}a_{ij}z_j\right) \\
&= Q\left(\eta(t_{n}, \alpha_{i}, b_i, c_i),\ y_{n},\ h\sum^{i-1}_{j=1}a_{ij}z_j,\ h\sum^{s}_{j=i+1}a_{ij}z_j\right) \label{eq:Q} \\
&= Q\left(\eta(t_{n}, \alpha_{i}, b_i, c_i),\ y_{n},\ h\sum^{i-1}_{j=1}b_{j}z_j\frac{a_{ij}}{b_j},\ h\sum^{s}_{j=i+1}b_{j}z_j\frac{a_{ij}}{b_j}\right) \label{eq:Q2} \\
&= I\left(\iota(t_{n}, \alpha_{i}, b_i, c_i),\ y_{n},\ hb_{1} z_1,\ \ldots,\ hb_{i-1}z_{i-1},\right. \nonumber\\
&\qquad\qquad\left. \  hb_{i+1} z_{i+1},\ \ldots,\ hb_{s}z_s\right) \\
&= I\left(\iota(t_{n}, \alpha_{i}, b_i, c_i),\ y_{n},\ e_1,\ \ldots,\ e_{i-1}, \ e_{i+1},\ \ldots,\ e_s\right).\label{eq:I}
\end{align}
At first, $h\sum^{s}_{j=1}a_{ij}z_{j}$ is divided into three parts, $h\sum^{i-1}_{j=1}a_{ij}z_j$, $ha_{ii}z_{i}$ and $h\sum^{s}_{j=i+1}a_{ij}z_j$ in Eq \eqref{eq:D3}. These three parts can be inputted separately. Their summation is approximated by the network.
As a result, the network is transformed to $G(\cdot)$ with the parameter $\beta$ and the inputs $y_n+ha_{ii}z_{i}$, $h\sum^{i-1}_{j=1}a_{ij}z_j$ and $h\sum^{s}_{j=i+1}a_{ij}z_j$. Then, we replace $z_i$ in $y_n+ha_{ii}z_{i}$ with $K(\cdot)$ according to Eq \eqref{eq:K}.
Thus, $y_n+ha_{ii}z_{i}$ can be approximated by a network $J(\cdot)$ with the parameter $\upsilon$ and the input $y_{n}$.
Consequently, $e_i$ can be approximated by a network with the parameter $\eta$ and the inputs $y_{n}$, $h\sum^{i-1}_{j=1}a_{ij}z_j$ and $h\sum^{s}_{j=i+1}a_{ij}z_j$. This network is written as $Q(\cdot)$ in Eq \eqref{eq:Q}. Afterwards, every $a_{ij}$ is adjusted to $b_{j}\frac{a_{ij}}{b_j}$ where $b_j \not=0$ in Eq \eqref{eq:Q2}.
Each $hb_{j} z_j$ can be inputted separately. Their weighted summation is approximated by the network.
As a result, $e_i$ can be approximated by a network with the parameter $\iota$ and the inputs $y_{n}$ and $hb_{j} z_j$ for $j=1,\ \ldots,\ s,\ j\not=i$. This network is written as $I(\cdot)$.
After replacing each $hb_{j} z_j$ with $e_j$ according to Eq \eqref{eq:ei},
$e_i$ is approximated by a network with the inputs $y_n$ and $e_j$ for $j=1,\ \ldots,\ s,\ j\not=i$.

Inspired by the Newton method used for the IRK methods, we consider that Eq \eqref{eq:E} can be used to offer the initial value of each $e_i$, which is written as $x_i$, as the input to Eq \eqref{eq:I}. i.e., Eq \eqref{eq:E} is rewritten as below.
\begin{equation}
x_i = E\left(\epsilon(t_{n}, \alpha_{i}, b_i, c_i),\ y_{n},\ x_1,\ \ldots,\ x_{i-1}\right). \label{eq:x} \end{equation} 
Next, we can apply $x_i$ to Eq \eqref{eq:I} in the following two ways:
\begin{equation}
e_i = I\left(\lambda(t_{n}, \alpha_{i}, b_i, c_i),\ y_{n},\ x_1,\ \ldots,\ x_{i-1}, \ x_{i+1},\ \ldots,\ x_s\right), \label{eq:not_replace}
\end{equation}
or
\begin{equation}
e_i = I\left(\phi(t_{n}, \alpha_{i}, b_i, c_i),\ y_{n},\ e_1,\ \ldots,\ e_{i-1}, \ x_{i+1},\ \ldots,\ x_s\right). \label{eq:replace}
\end{equation}
We use all the $x_j$ for $j=1,\ \ldots,\ s, j\not=i$ to compute $e_i$ in Eq \eqref{eq:not_replace}. However, we replace $x_j$ for $j=1,\ \ldots,\ i-1$ with the corresponding $e_j$ in Eq \eqref{eq:replace}. Correspondingly, the parameter of the network is changed from $\iota$ in Eq \eqref{eq:I} to $\lambda$ and $\phi$, respectively.

We consider that the clique block in CliqueNets is similar to the combination of Eq \eqref{eq:x} and Eq \eqref{eq:replace}. The clique block has two phases, referred to as Stage-I and Stage-II in the CliqueNet literature. To avoid confusion, these two phases are called Phase-I and Phase-II, respectively, in this paper. Phase-I is a dense block, so it is suitable to approximate Eq \eqref{eq:x} as mentioned above. The parts in the result of Phase-I are alternately updated in Phase-II. It is like the replacement for the input in Eq \eqref{eq:replace}.

Due to the advancement of CliqueNets, we transform the clique block to approximate $e_i$ in the RK block. The RK block is constructed as follows.

On the whole, an RK block is composed of a transformed clique block followed by a summation layer. The input of the RK block is $y_n$. The output of the transformed clique block is $y_n$ and all the increments $e_i$ for $i=1,\ \ldots,\ s$. The summation layer adds $y_n$ and $e_i$ for $i=1,\ \ldots,\ s$ to obtain $y_{n+1}$ according to Eq \eqref{eq:add_E}. The transformed clique block must conform to the following rules:

\textbf{Rule 1}: The number of channels of $y_n$ is restricted to the growth rate of the dense block in Phase-I. They are both written as $k$.

\textbf{Rule 2}:
The total times of growth in Phase-I are $s$, which is the number of stages of the RK methods. $s$ should be larger than 1 for updating alternately in Phase-II. $k$ channels outputted by every growth in Phase-I are $x_i$ for $i=1,\ \ldots,\ s$.

\textbf{Rule 3}:
In Phase-II, the updated $k$ channels for each $x_i$ are assumed to approximate $e_i$, representing the increment of each stage.

\textbf{Rule 4}
We introduce $y_n$ into Phase-II. In the original clique block, $y_n$ does not involve the computation in Phase-II directly. However, according to Eq \eqref{eq:replace}, $y_n$ should involve the computation of $e_i$ directly.

\textbf{Rule 5}
The weights are no longer shared between Phase-I and Phase-II.

\begin{figure*}[htbp]
\centering
\includegraphics[width=\textwidth]{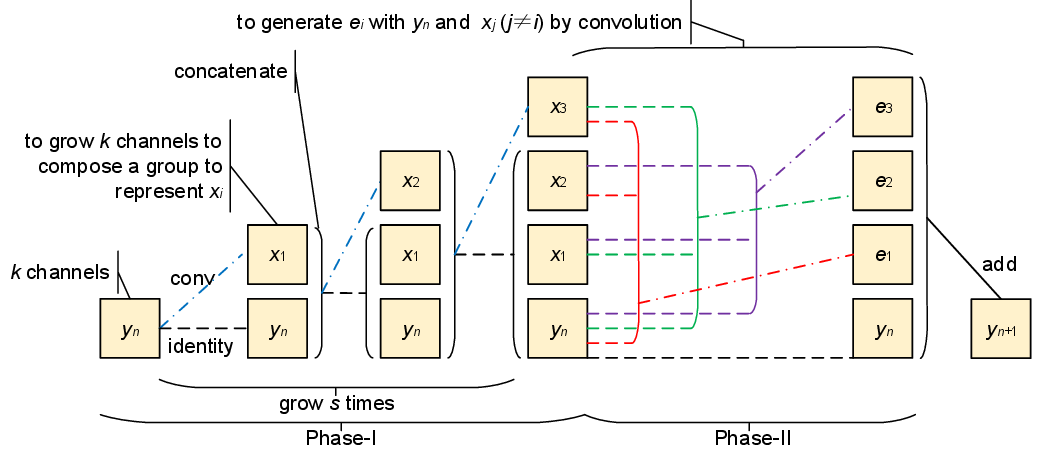}
\caption{Architecture of one time-step in an RKCNN-R using a 3-stage RK method. $y_n$ is the approximation of $y(t_n)$. A dense block grows $k$ channels every time to generate each $x_i$. After that, $y_n$ and every $x_j$ for $j=1,\ \ldots,\ 3,\ j\not=i$ are concatenated to generate the increment of each stage $e_i$ alternately. At last, $y_n$ and all the $e_i$ for $i=1,\ \ldots,\ 3$ are added to generate $y_{n+1}$ to complete a time-step.}
\label{fig:time-step}
\end{figure*}

\begin{table}[htbp]
\begin{center}
\begin{minipage}{0.9\textwidth}
\caption{A diagram of a time-step with five stages in an RKCNN-R. $y_n$ is deemed $x_0$. $w_{ij}^{(p)}$ for $i=0,\ldots,5,\ j=1,\ldots,5$ and $i\not=j$ is the weights of the parameter from $x_i$ to $x_j$ when $p=1$ or to $e_j$ when $p=2$. The superscript $p$ stands for Phase $p$ since the weights are not shared in RKCNNs. “\{\}” denotes the concatenation operator. The differences between the original clique block and the RK block in an RKCNN-R are represented in \textcolor{blue}{\bf{BLUE}}.}
\label{tb:structure}
\begin{tabular}{@{}llll@{}}
\toprule
Bottom Layers & Weights & Top Layer & Phase \\
\midrule
$y_n$ & $w_{01}^{(1)}$ & $x_1$ &  \\
$\{y_n, x_1\}$ & $\{w_{02}^{(1)},w_{12}^{(1)}\}$ & $x_2$ &  \\
$\{y_n, x_1, x_2\}$ & $\{w_{03}^{(1)},w_{13}^{(1)},w_{23}^{(1)}\}$ & $x_3$ & I \\
$\{y_n, x_1, x_2, x_3\}$ & $\{w_{04}^{(1)},w_{14}^{(1)},w_{24}^{(1)},w_{34}^{(1)}\}$ & $x_4$ &  \\
$\{y_n, x_1, x_2, x_3, x_4\}$ & $\{w_{05}^{(1)},w_{15}^{(1)},w_{25}^{(1)},w_{35}^{(1)},w_{45}^{(1)}\}$ & $x_5$ &  \\
\midrule
$\{$\textcolor{blue}{\boldmath$y_n$\unboldmath}$, x_2, x_3, x_4, x_5\}$ & $\{$\textcolor{blue}{\boldmath$w_{01}^{(2)},w_{21}^{(2)},w_{31}^{(2)},w_{41}^{(2)},w_{51}^{(2)}$\unboldmath}$\}$ & $e_1$ &  \\
$\{$\textcolor{blue}{\boldmath$y_n, x_1$\unboldmath}$, x_3, x_4, x_5\}$ & $\{$\textcolor{blue}{\boldmath$w_{02}^{(2)},w_{12}^{(2)},w_{32}^{(2)},w_{42}^{(2)},w_{52}^{(2)}$\unboldmath}$\}$ & $e_2$ &  \\
$\{$\textcolor{blue}{\boldmath$y_n, x_1, x_2$\unboldmath}$, x_4, x_5\}$ & $\{$\textcolor{blue}{\boldmath$w_{03}^{(2)},w_{13}^{(2)},w_{23}^{(2)},w_{43}^{(2)},w_{53}^{(2)}$\unboldmath}$\}$ & $e_3$ & II \\
$\{$\textcolor{blue}{\boldmath$y_n, x_1, x_2, x_3$\unboldmath}$, x_5\}$ & $\{$\textcolor{blue}{\boldmath$w_{04}^{(2)},w_{14}^{(2)},w_{24}^{(2)},w_{34}^{(2)},w_{54}^{(2)}$\unboldmath}$\}$ & $e_4$ &  \\
$\{$\textcolor{blue}{\boldmath$y_n, x_1, x_2, x_3, x_4$\unboldmath}$\}$ & $\{$\textcolor{blue}{\boldmath$w_{05}^{(2)},w_{15}^{(2)},w_{25}^{(2)},w_{35}^{(2)},w_{45}^{(2)}$\unboldmath}$\}$ & $e_5$ &  \\
\botrule
\end{tabular}
\end{minipage}
\end{center}
\end{table}
\begin{algorithm}
\caption{Calculate a time-step in RKCNN}\label{algo1}
\begin{algorithmic}[1]
\Require $y_{n} \vee s \vee \epsilon(t_{n}, \alpha_{i}, b_i, c_i) \vee \phi(t_{n}, \alpha_{i}, b_i, c_i) \vee \lambda(t_{n}, \alpha_{i}, b_i, c_i) \vee type$
\Ensure $y_{n+1}$
\State $i \Leftarrow 1$
\While{$i \leq s$}
    \State $x_i \Leftarrow E\left(\epsilon(t_{n}, \alpha_{i}, b_i, c_i),\ y_{n},\ x_1,\ \ldots,\ x_{i-1}\right)$
    \State $i \Leftarrow i + 1$
\EndWhile
\State $i \Leftarrow 1$
\While{$i \leq s$}
    \If{type is '-E'}\label{algln2}
        \State $e_i \Leftarrow x_i$
    \Else
        \If{type is '-I'}
                \State $e_i \Leftarrow I\left(\phi(t_{n}, \alpha_{i}, b_i, c_i),\ y_{n},\ e_1,\ \ldots,\ e_{i-1}, \ x_{i+1},\ \ldots,\ x_s\right)$
        \Else[type is '-R']
                \State $e_i \Leftarrow I\left(\lambda(t_{n}, \alpha_{i}, b_i, c_i),\ y_{n},\ x_1,\ \ldots,\ x_{i-1}, \ x_{i+1},\ \ldots,\ x_s\right)$
        \EndIf
    \EndIf
    \State $i \Leftarrow i + 1$
\EndWhile
\State $i \Leftarrow 1$
\State $y_{n+1} \Leftarrow y_{n}$
\While{$i \leq s$}
    \State $y_{n+1} \Leftarrow y_{n+1} + e_i$
    \State $i \Leftarrow i + 1$
\EndWhile
\end{algorithmic}
\end{algorithm}
\bigskip

So far, the combination of Eq \eqref{eq:x} and Eq \eqref{eq:replace} has been implemented in the RK block. We add the suffix "-I" to the name of RKCNNs to denote this structure. Additionally, we implement the combination of Eq \eqref{eq:x} and Eq \eqref{eq:not_replace} by removing the replacement in the alternate update in Phase-II. The suffix "-R" is added to the name of RKCNNs in order to denote the removal of the replacement. \figurename~\ref{fig:time-step} illustrates one time-step of an RKCNN-R using a 3-stage RK method as an example. We show the layers and the weights of each time-step in an RKCNN-R in Table~\ref{tb:structure} for comparison to the original clique block. Algorithm \ref{algo1} shows the pseudo-code for calculating a time step in RKCNN. The efficiency of RKCNN-E, RKCNN-I, and RKCNN-R are compared in Sect. \ref{sec:experiments}.

\section{Experiments}
\label{sec:experiments}

To evaluate the performance of RKCNNs on image classification tasks, experiments are conducted using the network architectures as follows. There is only one time-step in each period of the evaluated RKCNNs. In addition, the attentional transition, the bottleneck, and the multiscale feature strategy are adopted in RKCNNs, following CliqueNets. The attentional transition is a channel-wise attention mechanism in the transition layers. The 1$\times$1 bottleneck layers, which output $k$ channels to the following 3$\times$3 convolution layers, are used in the RK blocks. The multiscale feature strategy is a mechanism to collect the features of different map sizes into the final representation.

\subsection{Compared with the same order models}\label{subsec:comp_order}

We choose the ResNets with full pre-activation, RKNNs, RK-Nets, ODE-Nets, DenseNets, and CliqueNets for comparison. The former four models are numerical models, while the latter two models are the base of our RKCNNs. In this section, we aim to verify the performance of the proposed approximation to RK methods in RKCNNs. Hence, the comparison with other approximations must exclude the effects from the rest of the models and training scheme. Therefore, we put the various approximations into the unified framework and use the unified training scheme.

The pre-activation ResNets can be considered as adopting the RK methods of some unknown order. Most evaluated RKNNs, RK-Nets, and ODE-Nets adopted the 4th-order numerical methods. However, the IRK method adopted in RK-Nets is 5th-order since only this method is provided in the code\footnote{\url{https://github.com/rtqichen/torchdiffeq}} of \cite{chen2018neural}. Correspondingly, we evaluate the RKCNNs adopting the 2-stage methods, which may have 4th-order at most. Since the bottleneck is used in our RKCNNs, the pre-activation ResNets with bottleneck are added as part of the comparison. The bottleneck in the residual blocks retains its original structure \citep{he2016identity}, which is $\begin{bmatrix}\begin{smallmatrix}
1\times1,\ k/4 \\
3\times3,\ k/4 \\
1\times1,\ k
\end{smallmatrix}\end{bmatrix}.$
In addition, we ignore the time input in the ODE subnetwork of the RK-Nets to construct the ODE subnetwork of the RKNNs, which is time-invariant. The maximum number of iterations is four in ODE-Nets (4th-order implicit Adams method). All the evaluated numerical models have one time-step in each period except ODE-Nets since the adopted Adams methods are the LM method.

DenseNets and CliqueNets, which have the same number of layers as the evaluated RKCNNs, are used as baselines since RKCNN-E and RKCNN-I/R are constructed based on them, respectively. The bottleneck layers in the dense blocks and the clique blocks are the same as in the RK blocks, which are $\begin{bmatrix}\begin{smallmatrix}
1\times1,\ k \\
3\times3,\ k
\end{smallmatrix}\end{bmatrix}.$


\begin{table}[h]
\begin{center}
\begin{minipage}{\textwidth}
\caption{Classification errors evaluated on the test set of MNIST. The adopted method in ODE-Net or RK-Net is written in brackets after the model name. Each convolution in the preprocessors of these models outputs $k$ channels. The error is the lowest one over five runs with the mean$\pm$std in brackets. Results that outperform all the competing methods are {\bf BOLD} and the overall best result is represented in \textcolor{blue}{\bf BLUE}.}
\label{tb:MNIST}
\begin{tabularx}{\textwidth}{llXXXl}
\toprule
Model & $k$ & FLOPs (M) & Memory (MiB)\footnotemark[4] & Params (K) & Error (\%)  \\
\midrule
pre-act ResNet (bottleneck) & 32 & 7.19 & 767 & 34.86 & $0.44 (0.486\pm0.038)$ \\
pre-act ResNet & 32 & 8.43 & 767 & 52.23 & $0.33 (0.372\pm0.031)$ \\
\midrule
RKNN (4th-order ERK) & 32 & 12.44 & 767 & 52.36 & $0.35 (0.378\pm0.024)$ \\
RK-Net (4th-order ERK) & 32 & 12.60 & 767 & 52.94 & $0.33 (0.400\pm0.051)$ \\
RK-Net (5th-order IRK) & 32 & - & - & 52.94 & N/A\footnotemark[3] \\
ODE-Net (4th-order explicit) & 32 & 18.11 & 767 & 52.94 & $0.35 (0.380\pm0.028)$  \\
ODE-Net (4th-order implicit) & 32 & \textgreater19.49\footnotemark[2] & 767 & 52.94 & $0.34 (0.378\pm0.023)$  \\
\midrule
DenseNet ($T = 2$)\footnotemark[1] & 32 & 8.65 & 767 & 57.55 & $0.33 (0.358\pm0.036)$ \\
CliqueNet ($T = 2$)\footnotemark[1] & 32 & 10.27 & 767 & 57.55 & $0.30 (0.366\pm0.054)$ \\
\midrule
RKCNN-E-2 & 24 & 4.94 & 747 & 31.45 & $0.34 (0.384\pm0.032)$ \\
 & 30 & 7.63 & 767 & 48.85 & $0.33 (0.366\pm0.027)$ \\
 & 32 & 8.65 & 767 & 55.50 & $0.33 (0.364\pm0.039)$ \\
\midrule
RKCNN-I-2 & 20 & 4.10 & 747 & 31.01 & $0.33 (0.402\pm0.058)$ \\
 & 26 & 6.84 & 747 & 52.01 & $\mathbf{0.29(0.336\pm0.029)}$ \\
 & 32 & 10.27 & 767 & 78.41 & $\mathbf{0.29(0.316\pm0.033)}$ \\
\midrule
RKCNN-R-2 & 20 & 4.10 & 747 & 31.01 & $\mathbf{0.30(0.342\pm0.032)}$ \\
 & 26 & 6.84 & 747 & 52.01 & $\mathbf{0.29(0.336\pm0.057)}$ \\
 & 32 & 10.27 & 767 & 78.41 & \textcolor{blue}{$\mathbf{0.28(0.308\pm0.024)}$} \\
\botrule
\end{tabularx}
\footnotetext[1]{$T$ is the total number of layers in the period. We transplant the dense block and the clique block from the original models into the framework of RKCNNs.}
\footnotetext[2]{FLOPs for the different images are uncertain due to the iterations. However, it must carry out the starter, the 4th-order RK method with 3/8 rule, twice and the predictor-corrector pair once at least. Hence, "\textgreater" is used.}
\footnotetext[3]{This IRK method needs too much memory to train.}
\footnotetext[4]{It is the GPU memory cost by inferring only one image using Pytorch.}
\end{minipage}
\end{center}
\end{table}


\subsubsection{MNIST}\label{subsubsec:mnist}

We reproduce the pre-activation ResNet, ODE-Nets, and RK-Nets on MNIST, using the implementation provided by \cite{chen2018neural}. MNIST is a dataset of handwritten digits from 0 to 9. It has a training set of 60,000 images and a test set of 10,000 images. The image size is $28\times28$ pixels.

All the evaluated models adopt one period on MNIST following RK-Nets and ODE-Nets. Thus, there is no transition layer in them. The preprocessor and postprocessor use the ones in RK-Nets and ODE-Nets, but each convolution in the preprocessor outputs $k$ channels instead of fixed 64 channels. To be specific, the preprocessor is $\begin{bmatrix}\begin{smallmatrix}
3\times3,\ k  \\
4\times4,\ k,\ /2,\ 1 \\
4\times4,\ k,\ /2,\ 1
\end{smallmatrix}\end{bmatrix}.$
The RK blocks are filled in this framework to construct RKCNNs. In addition, we transplant the dense blocks and the clique blocks from the original models into this framework for comparison.

We follow all the training details of RK-Nets and ODE-Nets except the tolerance for ODE-Nets (4th-order implicit Adams method). The models are trained for 160 epochs using the mini-batch gradient descent (MGD) with a mini-batch size of 128. The learning rate is set to 0.1 initially and divided by 10 at 60, 100, and 140 epochs in the training procedure. A momentum of 0.9 is used. The random crop is applied to the training set. The tolerance 0.5 is adopted in ODE-Nets (4th-order implicit Adams method).

We compare the classification errors of the evaluated models on the test set of MNIST. The test data are shown in Table~\ref{tb:MNIST}. According to the experimental results, RKCNNs obtain the highest accuracy when all the models adopt $k=32$. In consideration of the number of FLOPs and parameters as well as the costed GPU memory, we shrink $k$ in RKCNNs to further compare. As a result, RKCNNs are more efficient than all the other models. The results are discussed in Sec. \ref{subsubsec:r}.

\subsubsection{SVHN and CIFAR}\label{subsubsec:svhn_cifar}

\begin{sidewaystable}
\sidewaystablefn
\begin{center}
\scalebox{0.7}{
\begin{minipage}{1.4\textheight}
\caption{Classification errors evaluated on the test sets of SVHN and CIFAR. $k_i$ is the number of the input channels in the $i$th period. The standard data augmentation is adopted on CIFAR. On SVHN, the data augmentation is not used but the dropout layers are added. The error is the lowest one over three runs with the mean$\pm$std in brackets. FLOPs and Params on SVHN are same as CIFAR-10 since they are both classified to 10 classes. Results that outperform all the competing methods are {\bf BOLD} and the overall best result is represented in \textcolor{blue}{\bf BLUE}.}
\label{tb:cifar}
\begin{tabular*}{1.4\textheight}{@{\extracolsep{\fill}}lllllllllll@{\extracolsep{\fill}}}
\toprule
& & & & & & \multicolumn{2}{@{}c@{}}{SVHN} & CIFAR-10 & \multicolumn{2}{@{}c@{}}{CIFAR-100} \\\cmidrule{7-8}\cmidrule{10-11}
Model & $k_1$ & $k_2$ & $k_3$ & {FLOPs(G)} & Mem(M)\footnotemark[6] & {Params(M)} & {Error(\%)} & {Error(\%)} & {Params(M)} & {Error(\%)} \\
\midrule
pre-act ResNet (bottleneck)\footnotemark[1] & 120 & 120 & 120 & 0.0845 & 871 & 0.113 & 2.78($2.83\pm0.07$) & 8.02($8.17\pm0.17$) & 0.145 & 31.03($31.04\pm0.02$) \\
pre-act ResNet\footnotemark[1] & 120 & 120 & 120 & 0.740 & 1227 & 0.845 & 1.76($1.83\pm0.06$) & 5.53($5.74\pm0.22$) & 0.878 & 24.78($25.08\pm0.26$) \\
\midrule
RKNN (4th-order ERK)\footnotemark[1] & 120 & 120 & 120 & 2.833 & 1231 & 0.846 & 1.67($1.71\pm0.04$) & 5.42($5.49\pm0.08$) & 0.879 & 24.91($25.23\pm0.37$) \\
RK-Net (4th-order ERK)\footnotemark[1] & 120 & 120 & 120 & 2.857 & 1235 & 0.853 & 1.70($1.72\pm0.03$) & 5.38($5.49\pm0.10$) & 0.885 & 25.04($25.43\pm0.57$) \\
RK-Net (5th-order IRK)\footnotemark[1] & 120 & 120 & 120 & -\ & - & 0.853 & N/A\footnotemark[5] & N/A\footnotemark[5] & 0.885 & N/A\footnotemark[5] \\
ODE-Net (4th-order explicit)\footnotemark[1] & 120 & 120 & 120 & 5.672 & 1237 & 0.853 & 1.73($1.75\pm0.02$) & 5.61($5.74\pm0.14$) & 0.885 & 26.71($33.21\pm6.10$) \\
ODE-Net (4th-order implicit)\footnotemark[1] & 120 & 120 & 120 & \textgreater6.375\footnotemark[3] & - & 0.853 & N/A\footnotemark[4] & N/A\footnotemark[4] & 0.885 & N/A\footnotemark[4] \\
\midrule
DenseNet ($T = 6$)\footnotemark[1]\footnotemark[2] & 120 & 120 & 120 & 0.930 & 1233 & 1.044 & 1.83($1.83\pm0.01$) & 4.92($4.95\pm0.05$) & 1.141 & 22.69($23.01\pm0.34$) \\
CliqueNet ($T = 6$)\footnotemark[1]\footnotemark[2] & 120 & 120 & 120 & 1.666 & 1235 & 1.061 & 1.73($1.83\pm0.10$) & 4.74($4.83\pm0.11$) & 1.158 & 23.00($23.13\pm0.15$) \\
\midrule
RKCNN-E-2\_2\_2 & 120 & 120 & 120 & 0.856 & 1231 & 0.977 & 1.80($1.83\pm0.03$) & 5.18($5.24\pm0.06$) & 1.009 & 23.76($24.26\pm0.50$) \\
\midrule
RKCNN-I-2\_2\_2 & 26 & 28 & 28 & 0.0842 & 773 & 0.103 & 2.40($2.60\pm0.20$) & 7.67($7.88\pm0.19$) & 0.111 & 30.94($31.57\pm0.56$) \\
 & 78 & 78 & 80 & 0.724 & 961 & 0.834 & \bf1.64($1.70\pm0.06$) & \bf4.71($4.79\pm0.14$) & 0.855 & \bf22.54($22.88\pm0.29$) \\
 & 120 & 120 & 120 & 1.707 & 1245 & 1.932 & \textcolor{blue}{\bf1.61($1.64\pm0.04$)} & \textcolor{blue}{\bf4.36($4.41\pm0.04$)} & 1.964 & \textcolor{blue}{\bf 20.43($20.84\pm0.36$)} \\
\midrule
RKCNN-R-2\_2\_2 & 26 & 28 & 28 & 0.0842 & 773 & 0.103 & 2.33($2.48\pm0.13$) & 7.63($7.77\pm0.12$) & 0.111 & 30.98($31.26\pm0.26$) \\
 & 78 & 78 & 80 & 0.724 & 963 & 0.834 & 1.72($1.74\pm0.02$) & 4.76($4.92\pm0.14$) & 0.855 & 22.80($22.87\pm0.08$) \\
 & 120 & 120 & 120 & 1.707 & 1245 & 1.932 & \bf1.62($1.65\pm0.02$) & \bf4.48($4.62\pm0.12$) & 1.964 & \bf20.69($20.89\pm0.20$) \\
\botrule
\end{tabular*}
\footnotetext[1]{We transplant the residual blocks, the dense blocks, the clique blocks or the ODE solvers from the original models into the framework of RKCNNs.}
\footnotetext[2]{$T$ is the total number of layers in three periods, $T/3$ per period.}
\footnotetext[3]{FLOPs for the different images are uncertain due to the iterations. However, it must carry out the starter, the 4th-order RK method with 3/8 rule, twice and the predictor-corrector pair once at least in each period. Hence, "\textgreater" is used.}
\footnotetext[4]{The model does not converge.}
\footnotetext[5]{This IRK method needs too much memory to train.}
\footnotetext[6]{It is the GPU memory cost by inferring only one image using Pytorch.}
\end{minipage}
}
\end{center}
\end{sidewaystable}


RKCNNs are also evaluated on SVHN and CIFAR. The SVHN dataset contains $32\times32$ colored digit images. There are 73,257 images in the training set, 26,032 images in the test set, and 531,131 images for additional training. The CIFAR-10 dataset contains 60,000 color images of size $32\times32$ in 10 classes, with 5,000 training images and 1,000 test images per class. The CIFAR-100 is similar to the CIFAR-10, except that it has 100 classes, each with 500 training images and 100 test images.

We adopt 3-period RKCNNs on SVHN and CIFAR following CliqueNets. The preprocessors, the transition layers, and the postprocessors in RKCNNs are the same as the ones in CliqueNets except that the number of channels outputted by the preprocessors in RKCNNs equals the growth rate in the first period but not 64 fixedly. Furthermore, the number of the output channels of the $1\times1$ convolution in the transition layers is not the same as the number of the input channels but equal to the growth rate in the next period.

In addition, for the purpose of comparing with the RKCNNs, we transplant the residual blocks of the pre-activation ResNets, the dense blocks of DenseNets, the clique blocks of CliqueNets, and the ODE solvers of the RKNNs, RK-Nets, and ODE-Nets into the framework of RKCNNs. Moreover, the final state of each period in the evaluated numerical models is outputted into both the adjacent transition layers and the postprocessor, just like RKCNNs. Nevertheless, the evaluated DenseNets and CliqueNets follow the strategy in their original models. To be specific, in the CliqueNets, the generated features in each period are outputted into the transition layers while these features and the input of each period are outputted into the postprocessor together. In the DenseNets, the generated features and the input of each period are all outputted into the transition layers. Additionally, we add the multiscale strategy to the DenseNets. In other words, all the features outputted into the transition layers are also outputted into the postprocessor.

On both SVHN and CIFAR, we follow all the training details of CliqueNets \citep{Yang_2018_CVPR} except the batch size and data augmentation. The weights of the convolution layer are initialized as done by \cite{He_2015_ICCV}. Moreover, the weights of the fully connected layer use Xavier initialization \citep{glorot2010understanding}. A weight decay of $10^{-4}$ and Nesterov momentum of 0.9 are used. The learning rate is set to 0.1 initially and divided by 10 at 50\% and 75\% of the training procedure. The tolerance 0.9 is adopted in ODE-Nets (4th-order implicit Adams method).

On SVHN, we use all the training samples without augmentation and divide the images by 255 for normalization following \cite{Yang_2018_CVPR}. We add a dropout layer \citep{JMLR:v15:srivastava14a} with a dropout rate of 0.2 after each convolution layer following \cite{Huang_2017_CVPR} and \cite{Yang_2018_CVPR}. The models are trained for 40 epochs using the MGD with a mini-batch size of 128.

On CIFAR, a standard data augmentation scheme is adopted following \cite{He_2016_CVPR}. The models are trained for 300 epochs using the MGD with a mini-batch size of 32.

We evaluate RKCNNs on SVHN and CIFAR to compare with the transplanted network models, which are initialized and trained in the same way as RKCNNs. The classification errors of the evaluated models on the test sets of SVHN and CIFAR are shown in Table~\ref{tb:cifar}. According to the experimental results, RKCNN-I-2\_2\_2 and RKCNN-R-2\_2\_2 obtain higher accuracies than all the other models on each dataset when $k=120$. In consideration of the number of FLOPs and parameters as well as the costed GPU memory, we reduce $k$ in RKCNNs for further comparison. As a result, RKCNN-I and RKCNN-R are more efficient than all the compared models. The results are discussed in Sec. \ref{subsubsec:r}.

\subsubsection{Results discussion}\label{subsubsec:r}

For the purpose of avoiding effects from the dimensionality reduction, we unify the dimensionality in each period of all the evaluated models firstly. To be specific, the models are set the same number of channels $k$ except for the same preprocessor, transition layers, and postprocessor. The pre-activation ResNets, RKNNs, RK-Nets and ODE-Nets win each other on MNIST, SVHN and CIFAR. The bottleneck does not bring benefit to the accuracies of the pre-activation ResNets on these datasets. RKCNNs obtain higher accuracy than other numerical models, except that RKCNN-E fails on SVHN. On every dataset, RKCNN-E obtains lower accuracy than RKCNN-I and RKCNN-R. The accuracies of RKCNN-I and RKCNN-R are similar to each other.

The 2-stage RKCNNs could reach 4th-order according to Sec. \ref{sec:RKCNNs}. The evaluated numerical models adopting the 4th-order RK methods should have a similar accuracy since there is only one time-step of size $h$ in each period and the local truncation error is $O(h^{5})$ (refer Sec. \ref{subsec:RKmethod}). In addition, the local truncation error of 4th-order LM methods is also $O(\Delta{t}^{5})$, where $\Delta{t}$ is time-step size \citep{butcher2008numerical}. For the evaluated ODE-Nets adopting LM methods, the period $h$ is divided into $r$ steps, i.e. $\Delta{t}=h/r$ where $r>1$. As a result, the local truncation error of ODE-Nets adopting 4th-order LM methods is $O(h^{5}/r^{5})$, which is lower than $O(h^{5})$. Hence, ODE-Nets adopting 4th-order LM methods should achieve higher accuracy than the models adopting 4th-order RK methods. The results against the theory are due to the different approximations of numerical methods since the rest of the evaluated numerical models are the same. Thus, RKCNN-I and RKCNN-R are better approximations of RK methods.

For a neural network model, the number of FLOPs and parameters as well as the costed GPU memory have to be considered. RKCNNs save the FLOPs but cost more parameters and memory than the evaluated numerical models with the same $k$. Therefore, we keep the stages unchanged and reduce the dimensionality of RKCNNs by shrinking the channels $k$ in order to decrease the costed parameters and memory. Although the errors of RKCNNs increase after reduction, the reduced RKCNNs are more efficient than not only numerical models but also DenseNets and CliqueNets. In order to verify the efficiency of RKCNNs further, we compare RKCNNs with the SOTA models in the next section.

\subsection{Compared with the SOTA models}

\begin{table}[h]
\begin{center}
\begin{minipage}{\textwidth}
\caption{Classification errors evaluated on the test set of MNIST. Each convolution in the preprocessors of RKCNN outputs $k$ channels. The error is the lowest one over five runs with the mean$\pm$std in brackets. Results that outperform all the competing methods are {\bf BOLD} and the overall best result is represented in \textcolor{blue}{\bf BLUE}.}
\label{tb:MNIST_sota}
\centering
\begin{tabular}{llll@{}}
\toprule
Model & $k$ & Params (M) & Error (\%)  \\
\midrule
RK-Net \citep{chen2018neural} & - & 0.22 & 0.47 \\
ODE-Net \citep{chen2018neural} & - & 0.22 & 0.42  \\
SONODE \citep{NEURIPS2020_418db2ea} & - & 0.28 & 0.36  \\
\midrule
RKCNN-E-2 & 32 & 0.06 & $\mathbf{0.33 (0.364\pm0.039)}$ \\
RKCNN-I-2 & 32 & 0.08 & $\mathbf{0.29(0.316\pm0.033)}$ \\
RKCNN-R-2 & 32 & 0.08 & \textcolor{blue}{$\mathbf{0.28(0.308\pm0.024)}$} \\
\botrule
\end{tabular}
\footnotetext{Note: The numbers of FLOPs are not reported in their papers, so they are not compared.}
\end{minipage}
\end{center}
\end{table}

\begin{sidewaystable}
\sidewaystablefn
\begin{center}
\scalebox{0.65}{
\begin{minipage}{1.5\textheight}
  \caption{Classification errors evaluated on the test sets of SVHN and CIFAR. $k_i$ is the number of the input channels in the $i$th period of RKCNN. FLOPs and Params on SVHN are same as CIFAR-10 since they are both classified to 10 classes. Results that outperform all the competing methods are {\bf BOLD} and the overall best result is represented in \textcolor{blue}{\bf BLUE}.}
  \label{tb:cifar_sota}
\begin{tabular*}{1.5\textheight}{@{\extracolsep{\fill}}lccccccccc@{\extracolsep{\fill}}}
\toprule
&  &  & & & \multicolumn{2}{@{}c@{}}{SVHN} & CIFAR-10 & \multicolumn{2}{@{}c@{}}{CIFAR-100} \\
\cmidrule{6-7}\cmidrule{9-10}
Model & {$k_1$} & {$k_2$} & {$k_3$} &\ FLOPs (G) & \ Params (M) & \ Error (\%) & \ Error (\%) & \ Params (M) & \ Error (\%)   \\
\midrule
pre-act ResNet \citep{he2016identity} & - & - & - & 4.71 & 10.2 & - & 4.49 & 10.2 & 22.46 \\
\midrule
DenseNet \citep{Huang_2017_CVPR} & - & - & - & 14.53 & 28.1 & 1.59 & 3.74 & 28.3 & 19.25 \\
 & - & - & - & 10.83 & 15.3 & 1.74 & 3.62 & 15.5 & 17.60 \\
 & - & - & - & 18.78 & 25.6 & - & 3.46 & 25.8 & 17.18 \\
\midrule
CliqueNet \citep{Yang_2018_CVPR} & - & - & - & 9.45 & 10.14 & 1.51 & - & - & - \\
 & - & - & - & 10.56 & 10.48 & 1.64 & - & - & - \\
\midrule
Momentum ResNet \citep{pmlr-v139-sander21a} & - & - & - & 14.79 & 139.0 & - & 4.76 & 139.2 & 23.2 \\
\midrule
RKCNN-E-5\_5\_5 & 80 & 80 & 80 & 1.05 & 1.19 & 1.60 & 4.43 & 1.21 & 21.69 \\
RKCNN-E-5\_5\_5 & 120 & 120 & 120 & 2.37 & 2.67 & 1.58 & 4.12 & 2.70 & 20.37 \\
\midrule
RKCNN-I-5\_5\_5 & 80 & 80 & 80 & 2.26 & 2.55 & \textcolor{blue}{\bf 1.50(1.55$\pm$0.05)} & 3.81 & 2.57 & 19.46 \\
RKCNN-I-5\_5\_5 & 120 & 120 & 120 & 5.07 & 5.72 & {1.52} & 3.55 & 5.75 & 18.41 \\
RKCNN-I-5\_5\_6 & 150 & 120 & 120 & 7.30 & 7.29 & {1.53} & \textcolor{blue}{\bf 3.38(3.55$\pm$0.15)} & 7.32 & 18.02 \\
\midrule
RKCNN-R-5\_5\_5 & 80 & 80 & 80 & 2.26 & 2.55 & 1.58 & 4.08 & 2.57 & 19.24 \\
RKCNN-R-5\_5\_5 & 120 & 120 & 120 & 5.07 & 5.72 & {\bf 1.51} & 3.60 & 5.75 & 18.24 \\
RKCNN-R-3\_4\_4 & 180 & 180 & 180 & 6.70 & 8.76 & {1.54} & {3.71} & 8.81 & \textcolor{blue}{\bf 17.00(17.46$\pm$0.42)} \\
\botrule
\end{tabular*}
\footnotetext{Note: We show the lowest test errors of three runs with mean$\pm$std for the best results of RKCNNs. The other test errors of RKCNNs are the random results.}
\end{minipage}
}
\end{center}
\end{sidewaystable}

We compare RKCNNs with the SOTA ODE-related models, the pre-activation ResNets, DenseNets and CliqueNets  on MNIST, SVHN and CIFAR. The architectures of RKCNNs and the training schemes on different datasets remain the same as what is in Sec.\ref{subsec:comp_order}. The test errors are shown in Table \ref{tb:MNIST_sota} and \ref{tb:cifar_sota}. According to the comparison, RKCNNs obtain higher accuracy than the competing models on MNIST. At the same time, the parameters of RKCNNs only account for about 21\textasciitilde36\% of the parameters of the competing models.

On SVHN and CIFAR, we increase the stages and channels in RKCNNs. On SVHN and CIFAR-10, RKCNN-I obtains higher accuracy than RKCNN-E, RKCNN-R and the competing models. On CIFAR-100, RKCNN-R obtains higher accuracy than RKCNN-E, RKCNN-I and the competing models. The parameters and FLOPs of RKCNNs are as low as 10\% of the parameters and FLOPs of competing models. Hence, RKCNN-I and RKCNN-R are more efficient.

\section{Conclusion}
\label{sec:conclusion}

From the dynamical systems view, we reinterpret the pre-activation ResNet and its variants which focus on improving the residual mapping. We consider that these models correspond to the RK methods but not only the forward Euler method.

We propose to employ the RK methods in moderate detail to construct the CNNs for the image classification tasks. The proposed network architecture can systematically generalize to the high order. It is referred to as RKCNN.

The experimental results demonstrate that RKCNNs surpass the state-of-the-art numerical models and their bases, i.e. the pre-activation ResNets, DenseNets, and CliqueNets, on MNIST, SVHN, and CIFAR.

With the help of the dynamical systems view and the various numerical ODE methods, including the RK methods, more neural networks can be constructed efficiently for the different tasks. Many aspects of RKCNNs and the dynamical systems view still require further investigation. For example, the network structure of RKCNNs may be improved to obtain higher efficiency. In addition, some other mathematical methods may also be approximated to construct the network models from the dynamical systems view. We hope that this work inspires future research directions.

\backmatter

\bmhead{Acknowledgments}
This work was supported by the Fundamental Research Funds for the Central Universities of China [No. N2024005-1].

%
\section*{Declarations}
\subsection*{Conflict of interest}

The authors declare that they have no conflict of interest.


\bibliography{RKCNN}

\begin{thebibliography}{28}
\providecommand{\natexlab}[1]{#1}
\providecommand{\url}[1]{{#1}}
\providecommand{\urlprefix}{URL }
\providecommand{\doi}[1]{\url{https://doi.org/#1}}
\providecommand{\eprint}[2][]{\url{#2}}
 \bibcommenthead

\bibitem[{Butcher(2008)}]{butcher2008numerical}
Butcher JC (2008) Numerical methods for ordinary differential equations. John
  Wiley \& Sons, The Atrium, Southern Gate, Chichester, West Sussex PO19 8SQ,
  England

\bibitem[{Chang et~al(2018{\natexlab{a}})Chang, Meng, Haber, Ruthotto, Begert,
  and Holtham}]{chang2018reversible}
Chang B, Meng L, Haber E, et~al (2018{\natexlab{a}}) Reversible architectures
  for arbitrarily deep residual neural networks. In: AAAI Conference on
  Artificial Intelligence

\bibitem[{Chang et~al(2018{\natexlab{b}})Chang, Meng, Haber, Tung, and
  Begert}]{chang2018multi}
Chang B, Meng L, Haber E, et~al (2018{\natexlab{b}}) Multi-level residual
  networks from dynamical systems view. In: International Conference on
  Learning Representations

\bibitem[{Chen et~al(2018)Chen, Rubanova, Bettencourt, and
  Duvenaud}]{chen2018neural}
Chen TQ, Rubanova Y, Bettencourt J, et~al (2018) Neural ordinary differential
  equations. In: Advances in Neural Information Processing Systems

\bibitem[{Dupont et~al(2019)Dupont, Doucet, and Teh}]{NEURIPS2019_21be9a4b}
Dupont E, Doucet A, Teh YW (2019) Augmented neural odes. In: Wallach H,
  Larochelle H, Beygelzimer A, et~al (eds) Advances in Neural Information
  Processing Systems, vol~32. Curran Associates, Inc.,
  \urlprefix\url{https://proceedings.neurips.cc/paper/2019/file/21be9a4bd4f81549a9d1d241981cec3c-Paper.pdf}

\bibitem[{E(2017)}]{E2017}
E W (2017) A proposal on machine learning via dynamical systems. Communications
  in Mathematics and Statistics 5(1):1--11. \doi{10.1007/s40304-017-0103-z},
  \urlprefix\url{https://doi.org/10.1007/s40304-017-0103-z}

\bibitem[{Gavahi et~al(2021)Gavahi, Abbaszadeh, and
  Moradkhani}]{GAVAHI2021115511}
Gavahi K, Abbaszadeh P, Moradkhani H (2021) Deepyield: A combined convolutional
  neural network with long short-term memory for crop yield forecasting. Expert
  Systems with Applications 184:115,511.
  \doi{https://doi.org/10.1016/j.eswa.2021.115511},
  \urlprefix\url{https://www.sciencedirect.com/science/article/pii/S0957417421009210}

\bibitem[{Glorot and Bengio(2010)}]{glorot2010understanding}
Glorot X, Bengio Y (2010) Understanding the difficulty of training deep
  feedforward neural networks. In: Proceedings of the thirteenth international
  conference on artificial intelligence and statistics, pp 249--256

\bibitem[{Haber et~al(2018)Haber, Ruthotto, Holtham, and
  Jun}]{haber2018learning}
Haber E, Ruthotto L, Holtham E, et~al (2018) Learning across
  scales---multiscale methods for convolution neural networks. In:
  Thirty-Second AAAI Conference on Artificial Intelligence

\bibitem[{He et~al(2015)He, Zhang, Ren, and Sun}]{He_2015_ICCV}
He K, Zhang X, Ren S, et~al (2015) Delving deep into rectifiers: Surpassing
  human-level performance on imagenet classification. In: The IEEE
  International Conference on Computer Vision (ICCV)

\bibitem[{He et~al(2016{\natexlab{a}})He, Zhang, Ren, and Sun}]{He_2016_CVPR}
He K, Zhang X, Ren S, et~al (2016{\natexlab{a}}) Deep residual learning for
  image recognition. In: The IEEE Conference on Computer Vision and Pattern
  Recognition (CVPR)

\bibitem[{He et~al(2016{\natexlab{b}})He, Zhang, Ren, and Sun}]{he2016identity}
He K, Zhang X, Ren S, et~al (2016{\natexlab{b}}) Identity mappings in deep
  residual networks. In: European Conference on Computer Vision, Springer, pp
  630--645

\bibitem[{Huang et~al(2017)Huang, Liu, van~der Maaten, and
  Weinberger}]{Huang_2017_CVPR}
Huang G, Liu Z, van~der Maaten L, et~al (2017) Densely connected convolutional
  networks. In: The IEEE Conference on Computer Vision and Pattern Recognition
  (CVPR)

\bibitem[{Krizhevsky(2009)}]{krizhevsky2009learning}
Krizhevsky A (2009) Learning multiple layers of features from tiny images.
  \urlprefix\url{https://www.cs.toronto.edu/~kriz/learning-features-2009-TR.pdf}

\bibitem[{Li et~al(2018)Li, Chen, Tai, and E}]{JMLR:v18:17-653}
Li Q, Chen L, Tai C, et~al (2018) Maximum principle based algorithms for deep
  learning. Journal of Machine Learning Research 18(165):1--29.
  \urlprefix\url{http://jmlr.org/papers/v18/17-653.html}

\bibitem[{Liao(2017)}]{liao2017towards}
Liao Q (2017) Towards more biologically plausible deep learning and visual
  processing. PhD thesis, Massachusetts Institute of Technology

\bibitem[{Lu et~al(2018)Lu, Zhong, Li, and Dong}]{pmlr-v80-lu18d}
Lu Y, Zhong A, Li Q, et~al (2018) Beyond finite layer neural networks: Bridging
  deep architectures and numerical differential equations. In: Dy J, Krause A
  (eds) Proceedings of the 35th International Conference on Machine Learning,
  Proceedings of Machine Learning Research, vol~80. PMLR, Stockholmsmässan,
  Stockholm Sweden, pp 3276--3285,
  \urlprefix\url{http://proceedings.mlr.press/v80/lu18d.html}

\bibitem[{Lu et~al(2020)Lu, Yu, Reddy, and Wang}]{lu2020few}
Lu Y, Yu F, Reddy MKK, et~al (2020) Few-shot scene-adaptive anomaly detection.
  In: European Conference on Computer Vision, Springer, pp 125--141

\bibitem[{Morales et~al(2021)Morales, Talavera, and
  Remeseiro}]{morales2021playing}
Morales D, Talavera E, Remeseiro B (2021) Playing to distraction: towards a
  robust training of cnn classifiers through visual explanation techniques.
  Neural Computing and Applications pp 1--13

\bibitem[{Netzer et~al(2011)Netzer, Wang, Coates, Bissacco, Wu, and Ng}]{37648}
Netzer Y, Wang T, Coates A, et~al (2011) Reading digits in natural images with
  unsupervised feature learning. In: NIPS Workshop on Deep Learning and
  Unsupervised Feature Learning 2011,
  \urlprefix\url{http://ufldl.stanford.edu/housenumbers/nips2011\_housenumbers.pdf}

\bibitem[{Norcliffe et~al(2020)Norcliffe, Bodnar, Day, Simidjievski, and
  Li\'{o}}]{NEURIPS2020_418db2ea}
Norcliffe A, Bodnar C, Day B, et~al (2020) On second order behaviour in
  augmented neural odes. In: Larochelle H, Ranzato M, Hadsell R, et~al (eds)
  Advances in Neural Information Processing Systems, vol~33. Curran Associates,
  Inc., pp 5911--5921,
  \urlprefix\url{https://proceedings.neurips.cc/paper/2020/file/418db2ea5d227a9ea8db8e5357ca2084-Paper.pdf}

\bibitem[{{\"O}ZT{\"U}RK et~al(2021){\"O}ZT{\"U}RK, Alhudhaif, and
  Polat}]{ozturk2021attention}
{\"O}ZT{\"U}RK {\c{S}}, Alhudhaif A, Polat K (2021) Attention-based end-to-end
  cnn framework for content-based x-ray image retrieval. Turkish Journal of
  Electrical Engineering \& Computer Sciences 29(SI-1):2680--2693

\bibitem[{Sander et~al(2021)Sander, Ablin, Blondel, and
  Peyr{\'e}}]{pmlr-v139-sander21a}
Sander ME, Ablin P, Blondel M, et~al (2021) Momentum residual neural networks.
  In: Meila M, Zhang T (eds) Proceedings of the 38th International Conference
  on Machine Learning, Proceedings of Machine Learning Research, vol 139. PMLR,
  pp 9276--9287,
  \urlprefix\url{https://proceedings.mlr.press/v139/sander21a.html}

\bibitem[{Süli and Mayers(2003)}]{endre2003an}
Süli E, Mayers DF (2003) An Introduction to Numerical Analysis. Cambridge
  University Press, The Edinburgh Building, Cambridge CB2 2RU, United Kingdom

\bibitem[{Srivastava et~al(2014)Srivastava, Hinton, Krizhevsky, Sutskever, and
  Salakhutdinov}]{JMLR:v15:srivastava14a}
Srivastava N, Hinton G, Krizhevsky A, et~al (2014) Dropout: A simple way to
  prevent neural networks from overfitting. Journal of Machine Learning
  Research 15:1929--1958.
  \urlprefix\url{http://jmlr.org/papers/v15/srivastava14a.html}

\bibitem[{Wang and Lin(1998)}]{wang1998runge}
Wang YJ, Lin CT (1998) Runge-kutta neural network for identification of
  dynamical systems in high accuracy. IEEE Transactions on Neural Networks
  9(2):294--307

\bibitem[{Yang et~al(2018)Yang, Zhong, Shen, and Lin}]{Yang_2018_CVPR}
Yang Y, Zhong Z, Shen T, et~al (2018) Convolutional neural networks with
  alternately updated clique. In: The IEEE Conference on Computer Vision and
  Pattern Recognition (CVPR)

\bibitem[{Şaban Öztürk(2021)}]{OZTURK2021102601}
Şaban Öztürk (2021) Class-driven content-based medical image retrieval using
  hash codes of deep features. Biomedical Signal Processing and Control
  68:102,601. \doi{https://doi.org/10.1016/j.bspc.2021.102601},
  \urlprefix\url{https://www.sciencedirect.com/science/article/pii/S1746809421001981}

\end{thebibliography}


\end{document}